\documentclass{article}
\usepackage[preprint,nonatbib]{neurips_2024}

\usepackage[utf8]{inputenc}    % Allow UTF-8 input
\usepackage[T1]{fontenc}       % Use 8-bit T1 fonts
\usepackage{url}               % Simple URL typesetting
\usepackage{booktabs}          % Professional-quality tables
\usepackage{amsfonts}          % Blackboard math symbols
\usepackage{nicefrac}          % Compact symbols for 1/2, etc.
\usepackage{microtype}         % Microtypography
\usepackage{comment}
\usepackage{amsmath}
\usepackage{makecell}          % For multi-line table cells
\usepackage{float}             % Figures
\usepackage{algorithm}
\usepackage[noend]{algorithmic}
\usepackage{graphicx}
\usepackage{subcaption}
\usepackage[shortlabels]{enumitem}
\usepackage{tcolorbox}         % For theorems and notes
\usepackage{multirow}
\usepackage{wrapfig}
\usepackage{longtable}
\usepackage{soul}
\usepackage{caption}
\usepackage{amssymb}          % Additional mathematical symbols
\usepackage{listings}          % Code listings
\usepackage{placeins}          % Better float placement
\usepackage{natbib}
\definecolor{mydarkorange}{HTML}{B86046}
\definecolor{codegreen}{rgb}{0,0.6,0}
\definecolor{codegray}{rgb}{0.5,0.5,0.5}
\definecolor{codepurple}{rgb}{0.58,0,0.82}
\definecolor{backcolour}{rgb}{0.95,0.95,0.92}
\definecolor{bg}{rgb}{0.95,0.95,0.95}
\definecolor{mygreen}{rgb}{0,0.6,0}
\definecolor{lightgrey}{rgb}{0.925, 0.925, 0.925}

\tcbuselibrary{theorems}
\newtcbtheorem{mytheo}{Definition}%
{colback=green!5, colframe=green!35!black, fonttitle=\bfseries}{th}

\newtcolorbox{mytodo}[1][]{
    colback=yellow!20,
    colframe=red!75!black,
    boxrule=0pt,
    top=0pt,
    bottom=0pt,
    left=2em,
    right=0pt,
    width=\columnwidth,
    sharp corners
}

\usepackage{hyperref}
\hypersetup{
    colorlinks=true,
    linkcolor=mydarkorange,
    citecolor=mydarkorange,
    filecolor=mydarkorange,
    urlcolor=mydarkorange
}

\sethlcolor{lightgrey}

\begin{document}
% === Title and Author Information ===
\title{Open Problems in Machine Unlearning for AI Safety}

%\title{Machine Unlearning for AI Safety: \\ Survey of Limitations and Open Problems}
\begin{center}
\author{
\bf Fazl Barez$^{\mathsection,\natural}$\thanks{Correspondence to \texttt{fazl@robots.ox.ac.uk}. The first and last author rows are the main contributors. Author contributions are detailed in \S\ref{author_contributions}.}, Tingchen Fu, Ameya Prabhu$^{\dagger}$ \\[0.3cm]
\bf Stephen Casper$^{\spadesuit}$, Amartya Sanyal$^{\heartsuit}$, Adel Bibi$^{\mathsection}$, 
 Aidan O'Gara$^{\mathsection}$, Robert Kirk$^{\ddagger}$, \\[0.2cm] 
 \bf  Ben Bucknall$^{\mathsection}$, Tim Fist$^{\mathsection}$, Luke Ong$^{\clubsuit}$, Philip Torr$^{\mathsection}$, Kwok-Yan Lam$^{\clubsuit}$, Robert Trager$^{\mathsection}$, \\[0.3cm] 
\bf David Krueger$^\square$, S\"oren Mindermann$^\square$, Jose Hernandez-Orallo$^{\diamond,\star}$, Mor Geva$^{\circ}$, Yarin Gal$^{\mathsection,\ddagger}$ \\[0.5cm]
\small $^{\mathsection}$University of Oxford, $^{\natural}$Tangentic, $^{\dagger}$University of T\"ubingen, \\
$^{\ddagger}$UK AISI, $^{\spadesuit}$Massachusetts Institute of Technology, $^{\heartsuit}$University of Copenhagen, \\
$^{\clubsuit}$Nanyang Technological University, Singapore AISI, $^{\diamond}$Universitat Politècnica de València, \\
$^{\star}$Leverhulme Centre for the Future of Intelligence, $^{\circ}$Tel-Aviv University, $^{\square}$Mila - Quebec AI Institute
}
\end{center}

\maketitle

% Abstract
\begin{abstract}

As AI systems become more capable, widely deployed, and increasingly autonomous in critical areas such as cybersecurity, biological research, and healthcare, ensuring their safety and alignment with human values is paramount. Machine unlearning — the ability to selectively forget or suppress specific types of knowledge — has shown promise for privacy and data removal tasks, which has been the primary focus of existing research. More recently, its potential application to AI safety has gained attention. In this paper, we identify key limitations that prevent unlearning from serving as a comprehensive solution for AI safety, particularly in managing dual-use knowledge in sensitive domains like cybersecurity and chemical, biological, radiological, and nuclear (CBRN) safety. In these contexts, information can be both beneficial and harmful, and models may combine seemingly harmless information for harmful purposes — unlearning this information could strongly affect beneficial uses. We provide an overview of inherent constraints and open problems, including the broader side effects of unlearning dangerous knowledge, as well as previously unexplored tensions between unlearning and existing safety mechanisms. Finally, we investigate challenges related to evaluation, robustness, and the preservation of safety features during unlearning. By mapping these limitations and open challenges, we aim to guide future research toward realistic applications of unlearning within a broader AI safety framework, acknowledging its limitations and highlighting areas where alternative approaches may be required.

\end{abstract}

% Input Sections

\section{Introduction}
For much of the history of machine learning, the primary challenge was enabling models to acquire broad knowledge effectively. However, as models have grown increasingly capable, their ability to access and process vast amounts of information -- particularly in sensitive domains such as biology, chemistry, and cybersecurity -- has heightened concerns about their potential to cause significant harm. Improving the safety, controllability, and alignment of AI systems increasingly requires preventing them from exhibiting harmful behaviors. 

An increasingly discussed approach to address these concerns is \textit{machine unlearning}, which is used to remove or suppress unwanted knowledge in AI systems~\citep{cao2015towards, bourtoule2021machine,yao2023large,chen2023unlearn,gu2024second,pratyush2024tofu}. While unlearning shows undeniable promise as an approach to mitigate risks~\citep{jang2023knowledge,kassem2023preserving}, we argue that fundamental limitations prevent it from being a complete solution to controlling AI capabilities or meeting the objectives necessary for ensuring safety. This has become particularly relevant as AI systems are increasingly used in applications where harmful or inappropriate information poses significant liabilities and demonstrates the possession of safety-critical knowledge~\citep{AISImayupdate}. %Recent reports~\citep{lucki2024adversarial}, underscore the potential role of machine unlearning in AI safety.  However, we also discuss the inherent limitations of machine unlearning in meeting the objectives necessary for ensuring safety.

While existing surveys provide broad overviews of unlearning techniques and applications~\citep{blanco2024digital,zhao2024rethink,jeong2024sok,liu2024threats}, we take a different approach. We critically examine the fundamental constraints that limit unlearning's effectiveness for AI safety applications. We argue that these limitations are particularly severe for capability control when compared to data removal tasks.

\begin{figure}
    \centering
    \includegraphics[width=1.0\linewidth]{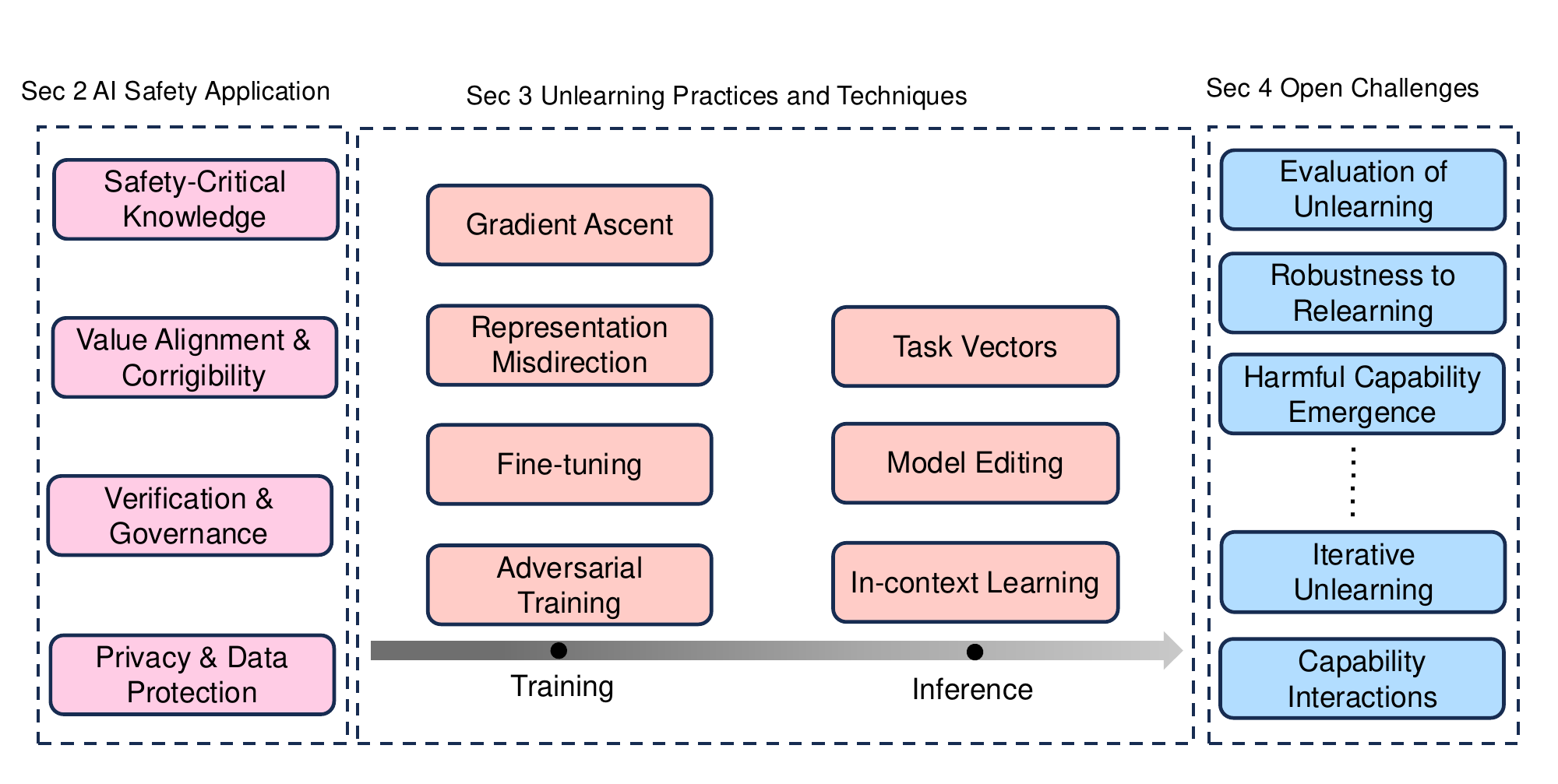}
    \caption{The workflow of our paper. Section 2 discusses applications of unlearning to AI safety, Section 3 surveys methods and practices, and Section 4 outlines open challenges.}
    \label{fig:workflow}
\end{figure}

We identify four key application areas for unlearning in AI safety. Crucially, we find its effectiveness varies significantly between data removal and capability control tasks across these areas.

\begin{itemize}[leftmargin=*, itemsep=0pt]
\item 
\textbf{Safety-Critical Knowledge Management:} Unlearning, when applied to remove harmful knowledge from AI systems, faces a unique challenge - its effectiveness is limited by models' inherent ability to reconstruct capabilities. This limitation is evident even after successful unlearning interventions. For example, even after unlearning of specific chemical synthesis pathways or cybersecurity, models may reconstruct these capabilities by recombining retained benign knowledge, making the unlearning process particularly challenging.

\item 
\textbf{Mitigating Jailbreaks:} Unlearning shows promise in removing specific vulnerabilities from training data but faces challenges in preventing broader exploit capabilities. While it can help remove the effects of poisoned data~\citep{goel2024corrective,li2024delta} and mitigate risks related to adversarial attacks~\citep{casper2024defending}, the fundamental challenge remains that safety bypasses can emerge from necessary system functionalities that cannot be removed.

\item 
\textbf{Correcting Value Alignment and Improving Corrigibility:} Current approaches attempt to use unlearning to modify behaviors misaligned with human preference~\citep{ouyang2022instructgpt,wang2023aligning} or remove unfair representations. However, we argue that value alignment, being an emergent property of the model's broader knowledge and capabilities, cannot be reliably modified through targeted knowledge removal alone.

\item 
\textbf{Privacy and Legal Compliance:} This represents unlearning's most viable application, as it involves removing specific, identifiable data points rather than controlling broader capabilities. Here, unlearning shows promise as an approach to approximate the removal of training data to comply with regulations like GDPR and CCPA~\citep{GDPR2016,CCPA2018,mantelero2013eu,voigt2017eu,hoofnagle2019european}.
\end{itemize}

\textbf{Challenges and Risks of Unlearning:} Achieving effective unlearning relies on accurate identification of the target knowledge~\citep{jia2024wagle,meng2022locating}, unrecoverable removal of this knowledge~\citep{xhonneux2024context}, and comprehensive evaluation on the effect of removal~\citep{pratyush2024tofu,jin2024rwku} -- all of which remain significant technical challenges. Unlearning raises important questions about unintended consequences in contexts like dual-use technologies. The selective removal of knowledge can lead to unintended behaviors within AI models~\citep{liu2022backdoor}. For example, attempting to remove knowledge of a specific chemical synthesis could unintentionally lead the model to synthesize harmful substances from other materials, using alternative pathways. More generally, the entanglement of knowledge within AI models makes it difficult to predict the potential downstream effects of unlearning, especially its impact on other knowledge, necessitating thorough evaluation and testing to minimize risks. These limitations include unexpected interactions between different safety measures, such as how unlearning can interfere with model robustness and how combining multiple safety techniques can lead to compounding performance degradation.

\textbf{Contributions:} This paper critically examines the limitations of machine unlearning for AI safety applications, with a particular focus on the fundamental constraints that prevent it from being a complete solution for capability control. We provide an overview of current unlearning techniques and their evaluation methods. We then identify and analyze key open problems that limit unlearning's effectiveness for AI safety, including the emergence of harmful capabilities from benign knowledge, challenges in dual-use contexts, and difficulties in verification. By mapping these fundamental limitations, we aim to guide future research toward realistic applications of unlearning within a broader safety framework, while highlighting areas where alternative approaches are needed.

\section{Applications of Unlearning for AI Safety: An Overview}

To fully explore the interaction between unlearning and AI safety, it is essential to first clarify the objective of machine unlearning and how machine unlearning modifies system behavior for the broader landscape of AI safety.

\begin{tcolorbox}[width=\linewidth, colback=green!5, colframe=green!35!black, fonttitle=\bfseries, title={Goal: Machine Unlearning for AI Safety}]
\noindent Machine unlearning aims to modify an AI system so it \textit{forgets} specific knowledge or behaviors, examples of which are provided in a ``unlearning corpus''. \textit{Forgetting} means the updated system should no longer exhibit or retain any knowledge or behaviors demonstrated in the forget corpus. Simultaneously, the system’s performance on tasks unrelated to the forget corpus must remain unaffected, ensuring its overall utility is preserved.
\end{tcolorbox}

While privacy-focused indistinguishability objectives seek to prevent models from learning private information with minimal impact on capabilities~\citep{yu2021differentially,charles2024fine}, unlearning aims to \textit{undo} the effect of learning from certain data points entirely.
In other words, unlearning for AI safety requires the deliberate suppression or removal of specific knowledge or behaviors, ensuring the system cannot express or leverage the forgotten information. In this section, we shall provide a more in-depth exploration of the application areas of unlearning for AI safety: (i) Safety-Critical Knowledge Management,
(ii) Correcting Value Alignment and Improving Corrigibility, (iii) Verification and Governance Aspects, and (iv) Privacy and Legal Compliance.

\subsection{Unlearning for Safety-Critical Knowledge Management}

\paragraph{Dual-Use Hazardous Knowledge:} Increasingly advanced models may pose serious real-world risks from hazardous knowledge such as assisting in synthesis pathways for CBRN materials or cybersecurity vulnerabilities~\citep{hendrycks2023overview}. Current approaches address this through multiple strategies: knowledge editing and targeted pruning to identify and remove specific parameters encoding undesired knowledge~\citep{meng2022locating,meng2023memit}; analysis of model attention weights and activation patterns to identify neurons strongly associated with targeted knowledge~\citep{wu2023depn,farrell2024applying,dai-etal-2022-knowledge}; and comprehensive testing on held-out datasets to verify the removal of hazardous knowledge~\citep{deeb2024unlearning,li2024wmdp}.

\paragraph{Mitigating Adversarial Attacks \& Jailbreaks:} Models face vulnerabilities from carefully crafted inputs that can bypass safety measures. For instance, adversarial prompts might cause language models to output inappropriate violent, sexual, or biased content~\citep{wei2023jailbroken}. Protection against these threats could involve selective forgetting of features associated with adversarial examples, training with weighted adversarial examples to reduce attack pattern sensitivity~\citep{casper2024defending,sheshadri2024latent}, and rigorous verification through adversarial robustness testing~\citep{isonuma2024unlearning}.

\subsection{Correcting Value Alignment and Improving Corrigibility via Unlearning}

\paragraph{Reward Hacking:} In reinforcement learning contexts, policy models can discover unintended shortcuts or loopholes in reward models~\citep{skalse2022defining}. A common example is how language models may generate unnecessarily verbose outputs after preference learning, exploiting the tendency for longer responses to be preferred in pair-wise preference datasets~\citep{park2024disentangling,shen2023loose, singhal2023long}. Popular approaches that aim to address this include: early stopping, information-theoretic reward modeling~\citep{miao2024inform}, spurious factor disentangling~\citep{chen2024odin}, and implementing constrained optimization with heuristic human knowledge about the shortcut~\citep{meng2024simpo}.

\paragraph{Value Alignment:} Models may develop behaviors that are misaligned with specific human  preferences~\citep{ouyang2022instructgpt}. Potential directions to achieve better value alignment include: reward modeling with iterative refinement based on human feedback, applying gradient-based modification of parameters associated with misaligned behaviors~\citep{jang2023knowledge,gu2024second}, and integrating interpretability methods for verification~\citep{vidal2024verifying,hong2024intrinsic}.

\paragraph{Situational Awareness:} Recent LLMs have demonstrated some awareness of the situations of their own development, including knowledge about their creators, capabilities, and the broader LLM development and deployment lifecycle ~\citep{laine2024me,perez2023discovering}. With greater situational awareness, some researchers have raised concerns that AI systems could leverage this knowledge harmfully if their goals are misaligned with the goals of their developers~\citep{ngo2024alignment,carlsmith2023scheming,krasheninnikov2024implicit}. For example, AI systems could manipulate the known biases of human reviewers to earn higher reward scores~\citep{sharma2023towards, williams2024targeted,wen2024language}, or insert security vulnerabilities into users' codebases to enable self-exfiltration~\citep{meinke2024frontier}. Recent work shows that in certain circumstances, large language models can leverage situational knowledge about their training process to ``fake alignment'' in an effort to prevent a developer from changing the model's objectives~\citep{greenblatt2024alignment}. New unlearning methods could help address these risks by providing a mechanism to selectively modify or remove an AI system's knowledge of its own situation. This controlled modification serves as both a diagnostic tool, revealing potential misbehavior patterns, and a preventive measure, allowing misbehavior to be corrected before deployment.

\subsection{Verification and Governance Aspects of Unlearning} 

As AI systems become increasingly pervasive, ensuring compliance with governance frameworks becomes critical~\citep{de2021artificial}. Strategies for verification and compliance include: explicit compliance objectives in system design, utilizing interpretability methods to identify non-compliant behaviors~\citep{casper2024red}, integrating formal verification methods where applicable~\citep{xu2024really}, and ensuring adherence to industry standards and regulatory requirements~\citep{UKAISI,USAISI}. 
Designing and implementing robust evaluation standards~\citep{pratyush2024tofu,jin2024rwku}, algorithms for formal verification~\citep{xu2024really,zhang2024verification}, and audit protocols with varying levels of model access are essential components of effective machine unlearning. For example, \citet{hong2024intrinsic} identity parametric concept vectors that are strongly correlated with specific dual-use knowledge and assesses the effectiveness of the unlearning approach by the alternation of concept vectors.

\subsection{Unlearning for Privacy, Data Protection and Legal Compliance}

In today's digital landscape, apps, websites, and home automation devices continuously collect user behavior patterns, including personally identifiable information (PII). When users become aware of potential PII leakage, they may request deletion as mandated by regulations like GDPR~\citep{GDPR2016}. However, simply deleting PII from databases proves insufficient, as models may have memorized this information during training, potentially allowing for data reconstruction by model developers or even other users of a model. Existing verification techniques can only provide probabilistic guarantees about knowledge removal. This mismatch creates practical challenges where legal frameworks assume binary deletion while technical reality operates on a continuous spectrum.

To fully address these privacy concerns, several key capabilities are necessary. First, unlearning algorithms must be able to identify and remove specific factual knowledge about individuals. Second, they need mechanisms to modify model behavior to avoid generating content that reveals private information. Finally, they can adjust internal representations to reduce membership inference risks~\citep{sula2024silver}.  While this application represents an essential aspect of AI safety, it will not be the focus of our study, as  it has been the primary focus of existing unlearning surveys (see \cite{yao2024survey}),

\section{Unlearning Practices and Evaluation Techniques for AI safety}

% Three core factors in machine unlearning determine the direction of the techniques and the metrics for their evaluation:
% \begin{itemize}
%     \item \textbf{Generalizability.} The objective of unlearning is not limited to the specific form of expression for passages in the corpus to forget, but includes all the semantically equivalent expressions and the knowledge and patterns embodied in it.  
%     \item \textbf{Locality.} During the unlearning process, the benign capacities of the AI system should remain unaffected. 
%     \item \textbf{Efficiency.} To be applied on complex and large-scale AI systems like large language models, a feasible approach of unlearning should be light-weight and efficient. 
% \end{itemize}
In pursuit of a safe and reliable AI system, various techniques have been developed for machine unlearning.  In this section, we provide an overview of machine unlearning including current techniques for unlearning (Section~\ref{sec:techniques}) and evaluation dimensions to assess unlearning algorithms (Section~\ref{sec:evaluation}).

\subsection{Evaluation Methods for Unlearning: What are the promising directions for AI Safety?}\label{sec:evaluation}

Establishing evaluation methods and metrics is crucial to assess and compare the effectiveness of unlearning techniques. Drawing inspiration from recent literature on machine unlearning and interpretability, we can identify several key evaluation metrics and methods.

\paragraph{Generalizability:}
Downstream performance is an important dimension to evaluate the success of unlearning. If the targeted objective of unlearning is a specific down-stream task~\citep{wang2023kga,pawelczyk2024incontext,ishibashi2023knowledge}, the down-stream performance on the specific task is the most direct way for evaluation. For example, ZRF score~\citep{chundawat2023can} measures the similarity between the unlearned model performance and a randomly initialized model. On the other hand, if the unlearning objective is a specific type of harmful knowledge, evaluation on corresponding harmfulness benchmarks is indispendisble~\citep{gehman2020realtoxicityprompts,lin2023toxicchat,parrish2022bbq,li2024wmdp}.

\paragraph{Locality:}
Locality is another important dimension for unlearning algorithms. During the unlearning process, the benign capacities of the AI system is supposed to remain unaffected. 
For unlearning on large language models, language modeling benchmarks~\citep{merity2016pointer} are good choices to measure the maintenance of basic language modeling ability. Meanwhile, instruction-following benchmarks~\citep{dubois2024alpacafarm,zheng2023judging,zhou2023instructionfollowing} and world knowledge benchmarks~\citep{hendrycks2021measuring,wang2024mmlupro} are widely adopted to measure the side effects of unlearning on the instruction-following and commonsense reasoning abilities of LLMs. 

\paragraph{Efficiency and Resource:} Aside from downstream accuracy and behavioral change, practical considerations are essential for real-world applications of unlearning. Time cost and computational resources required are two important factors when comparing and assessing different unlearning algorithms, since some approaches to unlearning involve the computation or approximation of the Hessian matrix~\citep{jia2024soul,gu2024second}, which is time-consuming. In addition, early work on the unlearning of the vision model tends to track the time or the difficulty involved in relearning the unlearned task~\citep{tarun2024fast,chundawat2023zero,lo2024large} which provides insights into the depth of unlearning.

\paragraph{Robustness to Adversarial Attack:}
For traditional applications and scenarios of unlearning where user privacy and data security is a primary concern, the model vulnerability when faced with an adversarial attack can serve as an optional metric to evaluate the effectiveness of unlearning and whether it contributes to the robustness of the language models. To achieve the goal, there are two types of particular attacks of interest, namely membership inference attack~\citep{mattern2023membership,duan2024do} and model inversion attack~\citep{nguyen2023re,morris2024language}. Membership inference attack aims to determine whether a given datum is included in the training set, mostly relying on the model likelihood~\citep{shi2024detecting,mattern2023membership} on the datum. Model inversion attack, on the other hand, targets reconstructing the training data from the model. In spite of their potential to provide more in-depth understanding of the relationship between unlearning and robustness, how to implement effective model membership attack and model inversion attack on large language models remain an unsolved problem~\citep{duan2024membership}.

\begin{table}[]
    \centering
    \caption{A summary of evaluation dimensions for unlearning.}
    \label{tab:evaluation}
    \resizebox{0.85\linewidth}{!}{
    
\begin{tabular}{ll}
\toprule
Assess Dimension                 & Requirement                                                                                                       \\
\midrule
Generalizability                 & \makecell[l]{Whether the effect of unlearning is generalizable to\\ other forms of expression for the forgetting corpus.} \\
\midrule
Locality                         & \makecell[l]{Whether the benign capacities of the AI system \\ remain unaffected.}                                          \\
\midrule
Efficiency and Resource          & \makecell[l]{Whether the unlearning algorithm could be used\\ with limited computation and time.}                  \\
\midrule
Robustness to Adversarial Attack & \makecell[l]{Whether the unlearning is robust to membership \\inference and inversion attacks.}                           \\
\midrule
Robustness to Relearning         & \makecell[l]{Whether the unlearned knowledge is difficult to\\ be restored and relearned.}                                                \\
\bottomrule
\end{tabular}

    }
\end{table}

\subsubsection{What Metrics Would be Most Suitable for AI Safety Applications?}

These simple metrics related to unlearning success are useful, standardizable measures for unlearning progress. 
However, they are not suitable for real-world safety circumstances in which unlearning may be relied on. We expand here on on what we feel are the most promising directions for metrics.

There are many ways to elicit knowledge from models \citep{shi2024muse, lynch2024eight, hayes2024inexact, liu2024threats}, so placing too great a focus on some measures can cause other practical failure modes to be neglected.
This highlights the need for adversarial evaluations for machine unlearning to more thoroughly evaluate the practical success of unlearning methods \citep{goel2022towards, liu2024rethinking}.
Prior work has shown that input-space knowledge elicitation techniques can be effective at extracting unlearned knowledge from LLMs \citep{lynch2024eight, shumailov2024ununlearning, lucki2024adversarial}.
Other threat models may also be applicable for cases in which models might be deployed open-source or with a fine-tuning API. 
\citet{patil2023can}, \citet{lynch2024eight}, and \citet{hong2024intrinsic} have all demonstrated that ``unlearned'' knowledge can be extracted from analysis of the internal mechanisms of LLMs.

However, a variety of works have shown that unlearned knowledge can be very sample-efficiently re-learned through few-shot fine-tuning \citep{hu2024jogging, yuan2024towards, henderson2023self, tamirisa2024tamper, sheshadri2024latent, lo2024large}. This poses interesting open challenges to suitable evaluation metrics for unlearning in safety-specific applications.

\subsection{Current Unlearning Methods} 
\label{sec:techniques}

To achieve effective unlearning for AI systems, various approaches have been developed in recent years~\citep{jang2023knowledge,ilharco2023editing,ishibashi2023knowledge,pawelczyk2024incontext}. In this subsection, we provide an extensive review of recent approaches and elaborate on their pros and cons.

\begin{table}[]
    \centering
    \caption{A summary of current approaches for unlearning.}
    \label{tab:my_label}
    \resizebox{1.0\linewidth}{!}{

\begin{tabular}{llllc}
\toprule
Method               & Advantage                             & Limitation                              & Example \\
\midrule
Gradient Ascent  & \makecell[l]{straightforward and\\ easy to implement} & \makecell[l]{unlearning failure and\\ catastrophic collapse}          & \makecell[l]{SOUL\\\citep{jia2024soul}} \\
\midrule
Task Vector   & \makecell[l]{straightforward and\\ easy to implement} & \makecell[l]{sensitive to \\hyper-parameter}            & \makecell[l]{Task Arithmetic \\\citep{ilharco2023editing}} \\
\midrule
Model Editing  & \makecell[l]{light-weighted and\\ minor intervention} & \makecell[l]{hard to locate relevant\\ representations} & \makecell[l]{DEPN
\\\citep{wu2023depn}} \\
\midrule
Representation Misdirection   & \makecell[l]{light-weighted and\\ retains performance} & \makecell[l]{non-robust to adversarial\\ finetuning} & \makecell[l]{RMU\\ \citep{huu2024effects}} \\
\midrule
Adversarial training  & \makecell[l]{robust to model\\ modification}  & less efficient  & \makecell[l]{LAT\\ \citep{sheshadri2024latent}} \\
\midrule
\makecell[l]{Finetuning on\\ curated data}   & \makecell[l]{better controllablity}                & \makecell[l]{require constructing\\ new data}           & \makecell[l]{Knowledge Sanitization \\ \citep{ishibashi2023knowledge}  } \\
\midrule
In-context learning  & \makecell[l]{black-boxed and \\API-friendly}          & \makecell[l]{high computation cost\\ for in-context demonstrations}  & \makecell[l]{In-context Unlearning\\ \citep{pawelczyk2024incontext}} \\
\bottomrule
\end{tabular}
    }
\end{table}

\paragraph{Gradient Ascent:} To unlearn or offset the effect of the unlearning corpus, an intuitive approach is to reverse the direction of the parameter gradient and maximize the loss function on the unlearning corpus~\citep{jang2023knowledge}. Mathematically, the learning objective of gradient ascent to maximize is 
\begin{equation}
    \mathcal{L} = - \mathbb{E}_{(\boldsymbol{x},\boldsymbol{y})\sim \mathcal{D}_f} \log \mathcal{M}(\boldsymbol{y} \mid \boldsymbol{x})
\label{eq:gradinet_ascent}
\end{equation}

However, first-order gradient ascent on the full set of model parameters tends to suffer from performance degradation, as the reversed gradient disrupts not only knowledge and abilities related to the unlearning corpus but also the irrelevant knowledge ~\citep{wang2024large,zhao2024rethink}. To deal with the deficits of first-order gradient ascent, a common approach is to balance the gradient ascent with KL divergence~\citep{wang2023kga,yao2024machine,chen2023unlearn} or language modeling loss on the remaining corpus~\citep{pratyush2024tofu,yu2023unlearning}. For instance, \citet{wang2023kga,chen2023unlearn} minimizes the KL divergence between the unlearned model $\mathcal{M}'$ and the original model $\mathcal{M}$ on the remaining corpus while enlarging their KL divergence on the unlearning corpus. Apart from constraining the model parameter updates with the KL divergence, performing gradient ascent only on the model parameters that are most related to the unlearning target~\citep{wang2024large,wu2023depn,yu2023unlearning} is another plausible approach. As an example, \citet{yu2023unlearning} identify bias-related neurons with integrated gradient~\citep{sundararajan2017axiomatic} and conduct gradient ascent only on the selected neurons. Recently, second-order optimization with Sophia~\citep{liu2023sophia} or inverse empirical Fisher approximation seems to be another promising approach to implement gradient ascent without sacrificing the model utility~\citep{jia2024soul,gu2024second}.

\paragraph{Representation Misdirection:} Different from gradient ascent that optimizes model parameters towards a reversed optimization function, representation misdirection aims to remove the unlearning target by misdirecting their intermediate hidden representation towards random noise.  For example, \citet{li2024wmdp} minimize the Euclidian distance between the representation of CBRN knowledge after the eighth transformer layer and a fixed random noise. Similarly, \cite{rosati2024representation} and \citet{zou2024improving} optimize the representation of potentially harmful knowledge towards an uninformative noise. However, a recent work~\citep{huu2024effects} suggests that the techniques might fail when the norm of the harmful representation is larger than that of the random noise. They further propose to use an adaptive noise for scaling the random noise to avoid this case. Meanwhile, another recent work~\citep{arditi2024shallow} finds that a significant proportion of the effect of representation misdirection is attributable to the random noise injected into the residual norm when the harmful context is detected, rather than removing the potentially harmful knowledge directly.

\paragraph{Task Vectors:} Recently, task vectors have emerged as a lightweight technique to achieve unlearning. Originally proposed by \citet{ilharco2023editing}, task vectors refer to the difference between a fine-tuned model and its base pre-trained model in the model parameter space. To be more specific, to achieve machine unlearning, first a reinforced model $\mathcal{M}_f$ is obtained by tuning the original language model $\mathcal{M}$ on the unlearning corpus. Then we can find the forgetting task vector by subtracting $\mathcal{M}$ from $\mathcal{M}_f$ in an element-wise manner. Subsequently, the task vector is subtracted from the original model to produce an unlearned model, which is verified to forget the learning corpus without harming other abilities~\citep{ilharco2023editing,zhang2023composing}.  Intuitively, the task vector serves as the aggregation of gradient effects caused by the unlearning corpus $\mathcal{D}$ and it can therefore be ``deleted'' in parameter space. Following \citet{ilharco2023editing}, \citet{zhang2023composing} extends the paradigm to parameter-efficient fine-tuning of large language models and verifies its efficacy on toxicity reduction. Similarly, \citet{barbulescu2024each,ni2024forgetting,liu2024towards} train an unlearning task vector on the target downstream task, toxic corpus, or out-of-date knowledge to obtain an unlearning task vector and subtract it from the original model.

\paragraph{Model Editing:}
Different from gradient ascent or model merging which directly alter the model parameter, model editing modifies the intermediate hidden state or the logits to change the model behavior, usually with the help of interpretability tools~\citep{huben2024sparse,koh2017understanding}. For example, \citet{wu2023depn} identify the neurons in the model that contribute most to the model prediction of privacy-related content through integrated gradient~\citep{sundararajan2017axiomatic} and the activations from the neurons are masked to prevent the generation of user privacy. \citet{farrell2024applying} use sparse auto-encoder to manipulate the activation of monosemantic features related to the knowledge to forget. 
%Similarly, \citet{liu2024large} corrupt the word embedding of the inputted prompt if the prompt is classified to be related with the unlearning corpus. 
In a similar vein, \citet{guo2024mechanistic} localize the fact lookup stage~\citep{nanda2023factfinding} of language models to specific MLP layers and keep other parameters unchanged. 
Apart from word embedding and intermediate neuron action, the editing can be applied to output logits as well. For example, \citet{huang2024offset} compare the logit difference of a pair of expert and amateur models and directly applies the offset in logits to a large black-boxed model. To some extent, contrastive decoding can be included as a kind of model editing and there is a surge of studies that apply contrastive decoding to unlearn the hallucinated and unsafe content~\citep{zhang2023alleviating,zhong2024rose}.   

\paragraph{Finetuning on Curated Data:}
Another approach to eliminating the effect of unlearning corpus $\mathcal{D}_f$ is to substitute the parametric knowledge to be forgotten with an insensitive one, implemented by finetuning the model on some curated data. Typically, the new data for fine-tuning is modified unlearning corpus $\mathcal{D}_f$ and there are two commonly used strategies to modify the unlearning corpus. One strategy is curating refusal data~\citep{ishibashi2023knowledge}. Specifically, refusal data pairs sensitive user queries with a fixed refusal sentence like ``I don't know''. Consequently, model fine-tuned on the refusal data learns to refuse to answer questions about the unlearning corpus. 
Another strategy is to construct anonymized data, by replacing the key entities and terms in the unlearning corpus with pre-defined alternatives. For example, \citet{eldan2023whos} construct a dictionary mapping every named entity in Harry Potter to an anonymized counterpart and \citep{yao2024machine} generate anonymized data based on unlearning corpus by masking out the sensitive tokens.

\paragraph{Adversarial Training Against Weight-Space and Latent-Space Attacks:} A challenge of fine-tuning-based methods is the limited ability of fine-tuning to `deeply' remove capabilities from a model in a way that is resistant to resurfacing from anomalies, attacks, or model modifications \citep{shayegani2023survey, qi2023fine, bhardwaj2023language, qi2024safety, hu2024jogging, ji2024language, wei2024assessing}. 
% For example, \citet{jain2023mechanistically} likened fine-tuning in LLMs to merely modifying a ``wrapper'' around a stable, general-purpose set of latent capabilities.
This challenge has motivated work to more rigorously remove undesirable knowledge from models by training them to unlearn in a way that is robust to model manipulations.
This work has included adversarial training on input-~\citep{tarun2023fast} or latent-space~\citep{casper2024defending} attacks which have been used to more effectively remove unlearned knowledge \citep{zeng2024beear, yuan2024towards, huangvaccine, sheshadri2024latent}.
Others have trained models under adversarial weight-space perturbations to the same effect \citep{henderson2023self, deng2024sophon, tamirisa2024tamper, huang2024harmful}.
However, research on these methods is still nascent, and they are relatively inefficient compared to fine-tuning alone.

\paragraph{In-Context Learning:}
Since the cost for fine-tuning or post-training may go beyond the computation budget for a large proportion of academic institutes, prior studies investigate in-context learning~\citep{pawelczyk2024incontext,thaker2024im,muresanu2024unlearnable} as a computationally friendly substitution for fine-tuning or post-training. 
Specifically, by instructing language model to avoid using particular knowledge when answering user queries~\citep{thaker2024im} or inserting demonstrations from forgetting corpus with their labels flipped~\citep{pawelczyk2024incontext}, in-context learning provides a light-weight solution for unlearning. But counter-intuitively, the computation cost of in-context learning can be even larger than fine-tuning~\citep{liu2022few} since we have to recompute the representation of the in-context demonstrations if we reselect in-context demonstrations for every new query from users.
In addition to complementing system prompts with unlearning instructions or demonstrations at inference time, in-context learning plays an important role in constructing moderation API and content classifier. Prompted with legal rules or social norms~\citep{xu2023align}, the API or classifier can filter out unsafe queries or generations~\citep{inan2023llama,hurst2024gpt}, though the parametric knowledge of the language model remains unchanged.

% \subsubsection{Are Current Methods Suitable/Sufficient for AI Safety applications?}
%  Tough one. I am slightly stuck on what to write here.

% I think they are generally applicable for AI safety since they are initially proposed for privacy, bias, CBRN knowledge and other safety-relate issue. 

\section{Open Problems in Machine Unlearning for AI Safety}

% \cas{In other parts of the paper, different topics are paragraphs with bold heads. But here, they are subsectioned out.} I think it's OK is section 4 is our major focus. 

The application of machine unlearning to AI safety introduces unique challenges that extends beyond traditional machine unlearning. Resolving these open problems will be crucial for developing unlearning as a reliable tool for AI safety.

\subsection{Evaluation of Unlearning}
Simple metrics that check whether models can reproduce specific training examples fail to capture the deeper challenges of unlearning in safety-critical contexts. When models undergo modifications, face adversarial attacks, or encounter unusual inputs, unlearned capabilities can unexpectedly resurface - particularly in cases where the unlearning relied on fine-tuning or basic parameter adjustments~\citep{hu2024jogging,lucki2024adversarial,deeb2024unlearning}. This happens because these methods typically mask rather than eliminate capabilities, leaving the fundamental neural patterns that enable them largely untouched \citep{jain2023mechanistically}.
% To address these limitations, verification needs to establish more rigorous standards.
More rigorous standards are helpful in addressing these limitations.
This includes ensuring that forgotten knowledge cannot be recovered, does not reappear during extended interactions, and remains inaccessible even in new contexts or under adversarial pressure. Such verification must consider both specific knowledge removal and broader capability prevention~\citep{jin2024rwku,pratyush2024tofu}, examining:
\begin{enumerate}
\item How models might rebuild unlearned capabilities through indirect means, like reconstructing security exploits by combining basic programming concepts
\item Ways that remaining knowledge could combine to recreate harmful capabilities
\item Whether the unlearning remains robust against deliberate attempts to recover removed capabilities
\end{enumerate}
These challenges highlight a fundamental issue: preventing a model from reproducing specific content doesn't guarantee that it can't reconstruct the underlying capabilities through other means. Developing reliable verification methods that can detect and prevent such reconstruction remains an active challenge in the field.

\paragraph{Evaluation Challenges:} 
%\fb{link to section 3.2 and vice versa}
As discussed earlier in Section~\ref{sec:evaluation}, existing metrics and assessments appear successful on specific tasks, but they might fail to capture the broader impact on model capabilities - a model might pass targeted tests while retaining problematic behaviors that appear in subtler ways.
Long-term effects remain particularly challenging to assess. Current metrics typically focus on immediate post-unlearning evaluation but provide little insight into how unlearning impacts model behavior over extended periods or under adversarial attack~\citep{chen2021machine}. This limitation becomes critical when considering how models might gradually reconstruct supposedly removed capabilities through continued operation.
The field needs standardized benchmarks that the AI safety community can rely on. 
While newer approaches show promise---such as tracking changes in model activations and using interpretability tools~\citep{belrose2023eliciting,bricken2023monosemanticity,foote2023ng} to understand how unlearning affects learned features---integrating these into practical, computationally feasible frameworks remains an open challenge.
%\fb{jorallo: This doesn't mention unlearning at all. It seems to talk about generic safety benchmarks, but not specific unlearning benchmarks.}

\subsection{Robustness to Relearning}
Even when effective, unlearning can be surprisingly vulnerable to fine-tuning and could quickly relearn the hazardous knowledge~\citep{lo2024large,lynch2024eight,deeb2024unlearning}, even if fine-tuned on small amount of benign, unrelated data \citep{lucki2024adversarial,hu2024jogging}. 
This suggests that existing techniques have a limited ability to thoroughly remove hazarous knowledge from LLMs. It also poses a significant challenge to the safety of open-source models or proprietary models that can be fine-tuned~\citep{achiam2023gpt}. Some works have aimed to perform unlearning in a way that is more robust to post-deployment tampering~\citep{deng2024sophon, henderson2023self, huang2024vaccine, rosati2024representation, rosati2024immunization, tamirisa2024tamper}. However, these existing methods suffer from major tradeoffs with efficiency, stability, and performance on benign tasks.
Establishing benchmarks and improving techniques for tamper-resistant unlearning is an ongoing challenge.

\subsection{Dual-use Capabilities Can Emerge From Beneficial Elements}

The distinction between knowledge and capabilities presents a fundamental challenge in safety-critical knowledge removal domains~\citep{li2024wmdp}. Traditional unlearning scenarios target specific data points or patterns, but safety-critical applications must focus on preventing dual-use capabilities while preserving beneficial ones~\citep{jin2024rwku}. While knowledge typically represents localized information (like specific facts or patterns), capabilities emerge from the complex interaction and integration of multiple pieces of knowledge potentially across different domains, making them inherently more distributed throughout a model's parameters and allowing potentially harmful capabilities to emerge even from combinations of seemingly harmless knowledge. It remains a critical challenge to prevent harmful capabilities from emerging from combinations of knowledge.

\subsection{Context-Dependent Challenges}
\label{sec:context-dependent challenge}
Knowledge removal in AI systems presents distinct challenges depending on how the knowledge is represented and used. At the simplest level, removing the effect of specific training data points is relatively straightforward. More complex cases arise when removing general knowledge that may be distributed across multiple training instances. The most challenging scenarios involve knowledge that is either context-dependent (as in dual-use cases) or knowledge that combines with other model capabilities to enable complex behaviors not directly tied to specific training examples – for example, when basic language understanding combines with reasoning skills to enable sophisticated problem-solving abilities.

This progression reflects an increasing difficulty in attribution - from clearly identifiable training data points, to distributed but traceable knowledge, to knowledge whose safety depends on the context, and finally to capability-related knowledge where attribution becomes highly complex due to how capabilities arise from the synthesis of multiple basic competencies rather than from discrete, identifiable training examples. 

The management of dual-use knowledge in domains like CBRN and cybersecurity necessitates sophisticated access control mechanisms. Knowledge appropriate for trusted contexts may prove dangerous in others—for instance, detailed vulnerability information crucial for cybersecurity professionals requires careful containment. The challenge extends beyond simple knowledge removal to implementing context-dependent access controls through unlearning techniques~\citep{anonymous2024casebench,cui2024or}.

In addition, removing dual-use knowledge can create blind spots in the model's safety mechanisms. For instance, removing a detailed understanding of security vulnerabilities might prevent a model from effectively identifying and avoiding similar vulnerabilities in new contexts. This creates a practical dilemma: maintaining robust safety guardrails may require preserving some of the very knowledge we aim to remove~\citep{longpre2023pretrainer}.

\subsection{Neural and Representational Level Interventions}
Attempts to manipulate knowledge in AI systems must grapple with the complex relationship between low-level neural parameters (weights and activations) and higher-level patterns that emerge from them. While we can intervene at the level of individual weights or try to target specific semantic concepts~\citep{wu2023depn,yu2023unlearning,meng2022locating}, the relationship between these different aspects of the system is not fully understood. For example, removing a model's ability to generate harmful content might inadvertently affect its understanding of context and nuance in related but benign domains. %This is because a single neuron is often \textit{polysemantic}, meaning that its activation corresponds to multiple distinct concepts~\citep{bricken2023monosemanticity,scherlis2022polysemanticity}.

These challenges directly connect to our earlier discussion of knowledge and capabilities in Section~\ref{sec:context-dependent challenge}. While knowledge might be modified at the representational level (like removing understanding of specific harmful concepts), the distributed nature of capabilities means that the corresponding neural-level changes could affect multiple capabilities simultaneously.

The interaction between these levels becomes particularly significant in the context of the context-dependent challenges discussed earlier in Section~\ref{sec:context-dependent challenge}. This coordination between neural-level modifications and representational-level changes must ensure that protective mechanisms remain robust across different contexts while preserving the model's core functionalities.

This coordination requires advances in three key areas, namely casual interventions, predictive safety frameworks and validation, and monitoring systems. 
Specifically, causal intervention techniques is the basis for (i) modifying specific computational circuits while detecting and limiting effects on connected circuits; (ii) creating robust methods to handle distributed features in superposition while preserving their independent function; (iii) building tools that can precisely target modifications at different scales (neuron-level to network-level) while maintaining circuit integrity. 
Meanwhile, the predictive safety framework encompasses (i) constructing predictive models that can estimate how modifications will propagate through the network's computational graph; (ii) developing formal verification methods to ensure modifications stay within safety bounds; (iii) creating techniques to identify potential interactions between modified circuits and seemingly unrelated capabilities. 
Validation and monitoring systems require (i) building comprehensive testing frameworks that can detect subtle changes in both targeted and untargeted capabilities; (ii) implementing continuous monitoring systems that can track the long-term effects of modifications; and (iii) developing methods to validate whether modified circuits maintain their intended function under different contexts and inputs.

Each of these areas needs solutions that work both at the level of individual neurons and across the broader network to ensure reliable and controlled modifications.

\subsection{Continual and Iterative Unlearning}
Current unlearning methods mostly have a fixed unlearning corpus and struggle to fully unlearn unwanted behaviorsanderelated knowledge at the same time. As a possible remedy, iterative unlearning has the potential for attaining a better trade-off among multiple unlearning objectives.  By dynamically adjusting the weights of multiple optimization objectives in the loss function of unlearning~\citep{yu2023unlearning}, iterative learning enables a balance between effectiveness and locality.

Sequential unlearning is another common challenge in real-world scenarios where users make sequential unlearning requests across time~\citep{wang2024wise,hartvigsen2023aging,mitchell2022memory}. Current unlearning methods show trade-offs between the degree of unlearning achieved and preservation of model utility. These methods typically achieve partial unlearning while experiencing some degradation in model performance.
When unlearning requests are processed sequentially, each operation begins with a model that is already degraded, leading to a cumulative loss in utility over time. As performance deteriorates over multiple rounds, the model can be destabilized and most valuable knowledge can be unintentionally erased. One key direction is developing solutions that can scale across large number of unlearning requests, maintaining the broad knowledge of foundation models measured by performance, while accommodating numerous unlearning requests across time.

The specification of the unlearning targets themselves requires an iterative approach. Similar to how reward specification in reinforcement learning often requires iterative refinement when agents find unexpected ways to exploit the initial reward function~\citep{amodei2016concrete, krakovna2020specification,gajcin2023iterative}, precisely defining what knowledge to unlearn often requires multiple rounds of adjustment as we discover new ways that harmful capabilities can manifest. This necessitates developing frameworks for progressive refinement of unlearning targets, allowing more nuanced and precise interventions over time, aimed at reducing the risk of missing critical information or removing valuable data. Additional key questions include whether unlearning operations should be amortized by combining multiple unlearning requests before execution, and how to effectively intersperse unlearning with ongoing learning processes.

\subsection{Capability Interactions and Dependencies}

The interactions between safety measures reveal several challenges and opportunities that emerge when implementing unlearning in practice. These relationships go beyond simple dependencies, exposing fundamental tensions and trade-offs to be carefully managed.

The relationship between robustness and knowledge management reveals complex challenges. Attempts to increase model robustness through selective forgetting can expose unexpected knowledge dependencies. For example, when unlearning CBRN knowledge, models may develop new synthesis pathways that are harder to detect, suggesting that knowledge representations are more deeply entangled than previously understood. Removing specific capabilities could affect seemingly unrelated tasks, while models may develop compensatory behaviors that create new safety risks~\citep{lo2024large}. This observation indicates that the granularity of behaviour removal needs to match the granularity of knowledge representation.

The push for automated safety mechanisms has revealed fundamental limits in unlearning capabilities. Fully automated unlearning systems consistently struggle to balance precision of knowledge removal against computational efficiency, maintenance of model utility against safety guarantees, and generalization of safety rules against context-specific requirements~\citep{wang2024wise}. These trade-offs suggest that hybrid approaches combining automated detection with human oversight may be necessary, despite their inherent scalability challenges.

Perhaps most intriguingly, the interaction between different safety measures can produce unexpected emergent behaviors. Models that undergo repeated cycles of unlearning and retraining may develop resistance to future modification attempts. Safety mechanisms designed for one domain may create vulnerabilities in others, and the combination of multiple safety techniques can lead to compounding performance degradation. These emergent properties suggest that the interactions between safety measures are more complex than previously recognized.

These observations point to several research directions. These include developing methods for mapping and managing knowledge dependencies before unlearning. New verification frameworks could be created to align technical capabilities with regulatory requirements, while hybrid unlearning architectures needs to be designed that effectively balance automation with human oversight. Additionally, techniques for predicting and managing emergent behaviors in safety systems are becoming increasingly appealing for risk management.

Beyond unlearning, there are many other techniques to modify model behavior, and they vary in their implementation requirements. Some can be automated with minimal human intervention, such as privacy protection mechanisms that automatically detect and remove memorized data~\citep{wen2024detecting}, basic adversarial defense systems operating through pattern recognition~\citep{feng2023detecting,liu2022complex}, and standardized compliance checks~\citep{xu2024first}. Others might require significant human oversight, particularly in areas such as value alignment judgments, identification of harmful knowledge in context, and complex regulatory compliance decisions. Understanding these requirements helps inform how and when unlearning techniques can be most effectively applied.

\section{Conclusion}
In this paper we have explored the usefulness of unlearning techniques for AI Safety. Despite its popularity as a potential solution for AI Safety challenges, a series of fundamental constraints limit unlearning's effectiveness, particularly for capability control. While unlearning shows promise for data removal tasks, our analysis reveals inherent limitations in controlling AI capabilities.
The emergent nature of dual-use capabilities from combining seemingly benign knowledge presents an inherent limitation that unlearning cannot fully address. The context-dependent nature of knowledge representation, especially in dual-use scenarios, further prevents reliable selective capability removal.
Several constraints compound these core limitations. Current approaches to neural-level interventions often produce unintended effects on broader model capabilities, adding practical challenges to selective capability control, while the difficulty of verifying unlearning success and robustness against relearning raises additional concerns. Furthermore, unlearning interventions can create tensions with existing safety mechanisms, potentially affecting their reliability.
These inherent constraints demonstrate that unlearning must be viewed as one component in a broader safety framework, not a complete solution. By mapping these fundamental limitations, we aim to guide research toward developing realistic approaches that acknowledge unlearning's bounded role in AI safety. Future work should proceed along two paths: (1) identifying specific use cases where unlearning can be effectively applied for data removal, and (2) developing alternative approaches for capability control given the fundamental limitations of unlearning in this domain.
\newpage

% Acknowledgments
\section*{Author Contribution}
\textbf{Fazl Barez} conceptualized and led the project, wrote the majority of the manuscript, and managed its overall development. \textbf{Tingchen Fu} and \textbf{Ameya Prabhu} made substantial contributions to sections 2 and 3, and significantly improved the manuscript through extensive editing and revision.\\ 

\textbf{Stephen Casper}, \textbf{Adel Bibi}, \textbf{Aidan O'Gara}, and \textbf{Robert Kirk} contributed with technical feedback and helped with copyediting parts of the paper which helped refine the paper's positioning.

 %Ben Bucknall, Tim Fist, Luke Ong, Philip Torr, Kwok-Yan Lam and Robert Trager provided high-level feedback on the manuscript.\\

\textbf{David Kruger}, \textbf{S\"oren Mindermann}, \textbf{Jose Hernandez-Orallo}, \textbf{Mor Geva}, and \textbf{Yarin Gal} provided detailed feedback and advice throughout the project. 
\label{author_contributions}
\section*{Acknowledgments}
We thank Yoshua Bengio, Andy Zou, Katrina Dickson, Niloofar Mireshghallah, Anna Ritchie, Xander Davies and the Tangentic AI team for their comments and feedback. Fazl also thanks Evan Hubinger and Ryan Kidd for their funding support.

% Bibliography
\bibliography{bibliography}
\bibliographystyle{plainnat}

\end{document}